\documentclass{article}

\usepackage[utf8]{inputenc}
\usepackage[T1]{fontenc}
\usepackage{amsmath,amssymb,amsfonts}
\usepackage{graphicx}
\usepackage{booktabs}
\usepackage{hyperref}
\usepackage{url}
\usepackage{authblk}       
\usepackage{natbib}        
\usepackage{adjustbox}     
\usepackage{multirow}      
\usepackage{caption}       
\usepackage{subcaption}    
\usepackage{xcolor}        
\usepackage{float} 
\usepackage{xurl}
\usepackage{array}    
\usepackage{ragged2e} 

\usepackage{cleveref}

\usepackage{colortbl}       
\usepackage{pifont}         
\usepackage{cuted}

\newcommand{\qmark}{\textcolor{green!70!black}{\ding{51}}}%
\newcommand{\xmark}{\textcolor{red!70!black}{\ding{55}}}%

\usepackage{amsmath,amsfonts,bm}

\def\eqref#1{equation~\ref{#1}}

\def\1{\bm{1}}

\def\vtheta{{\bm{\theta}}}

\def\vh{{\bm{h}}}

\DeclareMathAlphabet{\mathsfit}{\encodingdefault}{\sfdefault}{m}{sl}
\SetMathAlphabet{\mathsfit}{bold}{\encodingdefault}{\sfdefault}{bx}{n}

\def\methodFamilyName{\texttt{Superscopes}}

\def\sourceSeqLen{n}
\def\sourcePosition{i}
\def\sourcePrompt{S}
\def\sourceLayer{\ell}
\def\numSourceLayers{L}
\def\sourceModel{\mathcal{M}}
\def\sourceHidden{\vh}
\def\sourceHiddenDim{d}

\def\targetSeqLen{m}
\def\targetPosition{i^*}
\def\targetPrompt{T}
\def\targetLayer{\ell^*}

\def\targetHidden{\bar{\vh}}

\def\sourceMLP{mlp}
\def\sourcePreMLP{\vh_\text{PreMLP}}
\def\targetMLP{mlp^+}

\def\transformation{f}

\definecolor{myblue}{RGB}{218, 232, 252}
\definecolor{myred}{HTML}{F8CECC} 
\definecolor{mygreen}{HTML}{D5E8D4} 
\definecolor{myorange}{HTML}{C35500}

 \usepackage[accepted]{icml2024}

\title{\textbf{Superscopes: Amplifying Internal Feature Representations \\
for Language Model Interpretation}}
\author[*]{Jonathan Jacobi}
\author[*]{Gal Niv}

\affil[*]{{ Independent Researchers}}

\begin{document}
\maketitle

\begin{strip} 
    \centering
    \includegraphics[width=0.7\textwidth, keepaspectratio]{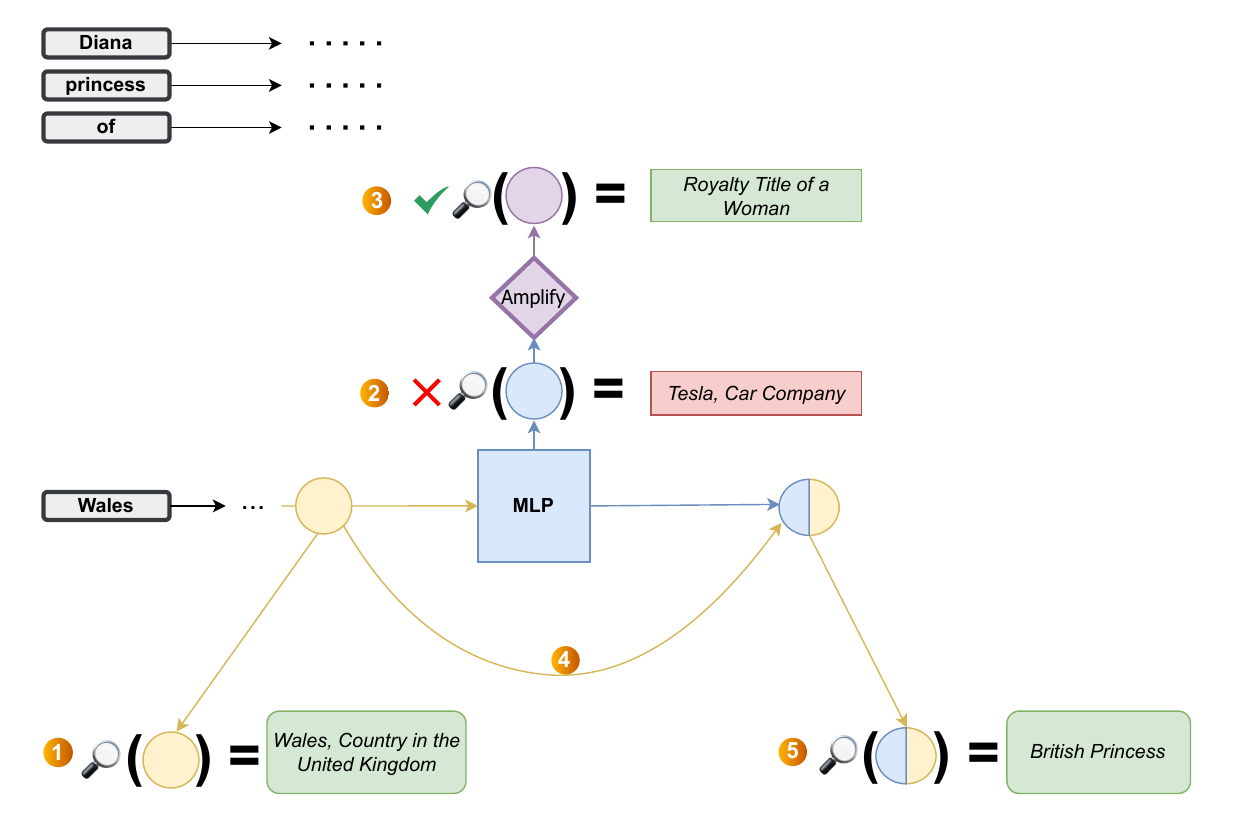}
     \captionof{figure}{Using \methodFamilyName{} to interpret a hidden state of the token "Wales" in an advanced layer. 
    \textcolor{myorange}{\textbf{Step 1:}} Using \emph{Patchscopes} (\emph{magnifying glass}) to extract the residual stream's meaning before applying the MLP, a sensible explanation is extracted. 
    \textcolor{myorange}{\textbf{Step 2:}} Applying Patchscopes to inspect the meaning of the MLP output, yields a nonsensical explanation ("Tesla, Car Company"). 
    \textcolor{myorange}{\textbf{Step 3:}} Amplifying the MLP output (\emph{"Superscoping"}) and inspecting it using Patchscopes yields \emph{\textbf{"Royalty Title of a Woman"}} - a logical explanation given the examined prompt. 
    \textcolor{myorange}{\textbf{Step 4:}} The transformer adds the MLP output to the residual stream. 
    \textcolor{myorange}{\textbf{Step 5:}} Using Patchscopes to interpret the new residual stream reveals that the resulting hidden state encodes the concept \emph{"British Princess"}. 
    This demonstrates how:
    \textbf{"Wales" + "Royalty Title of a Woman" = "British Princess"}, thereby clarifying the model’s inner contextualization and thinking process.}
    \label{fig:mainfig}
\end{strip}

\begin{abstract}
Understanding and interpreting the internal representations of large language models (LLMs) remains an open challenge. Patchscopes (\citet{ghandeharioun2024patchscope}) introduced a method for probing internal activations by patching them into new prompts, prompting models to self-explain their hidden representations. We introduce \methodFamilyName{}, a technique that systematically amplifies superposed features in MLP outputs (multilayer perceptron) and hidden states before patching them into new contexts. Inspired by the \emph{``features as directions''} perspective (\citet{elhage2022superposition}) and the \emph{Classifier-Free Guidance (CFG)} approach from diffusion models, \methodFamilyName{} amplifies weak but meaningful features, enabling the interpretation of internal representations that previous methods failed to explain—all without requiring additional training. This approach provides new insights into how LLMs build context and represent complex concepts, further advancing mechanistic interpretability.
\end{abstract}

\section{Introduction} 
\label{sec:intro}

Over the past few years, large language models (LLMs) have been researched and advanced significantly, leading to technological advancements that previously seemed impossible. However, due to the black-box nature of the models, there has been a lot of focus (\citet{casper2022sok, madsen2022post, patel2021mapping, nanda-etal-2023-emergent}) on understanding the inner-workings of the models and achieving greater clarity on their internal decision-making process.

Interpretation of a model's inner workings proved to be critical for various reasons, such as AI alignment and safety (\citet{bereska2024mechanisticinterpretabilityaisafety}), and understanding the underlying logic behind a model's reasoning (\citet{arkoudas2023gpt4cantreason}).
Many different methods to tackle interpretability were invented over the past few years, such as training-based interpretability methods that train linear classifiers (probes) from internal hidden states (\citet{belinkov2019analysis, belinkov2022probing}), projecting hidden states to the vocabulary space (\citet{logitlens, belrose2023eliciting}), and interfering mid-computation to identify hidden states with significant effect on predictions (\citet{meng2022locating,singh-etal-2020-bertnesia,wang2022interpretability,geva-etal-2023-dissecting}).

Although these methods were successful in showing great progress, research continued and researchers came up with a new novel idea of \emph{models self-interpreting their internal representations}. The core concept behind self-interpreting is that different parts of models can be given as an \textit{input} to the models in certain ways that would allow translating the components to natural language.

A key piece of research in the field is \textit{Patchscopes} (\citet{ghandeharioun2024patchscope}), introducing a framework for patching hidden states from different layers of a source prompt into a target prompt, enabling the extraction of meaning from the patched-in hidden states. Another significant research is \textit{Speaking Probes} (\citet{dar2023speakingprobes}), which introduced the idea of patching and interpreting the model's weights, specifically the feed-forward keys, which shows meaning behind certain parameters of the model. An additional approach taken was \textit{SelfIE} (\citet{chen2023selfie}), which introduced ways to also leverage language-based interpretability method in various different applications.

Building on the concepts behind Patchscopes, Speaking Probes, and SelfIE (we call those \emph{\textbf{"Self Interpreting Methods"}}), it is natural to apply these techniques to the outputs of the MLP and Attention modules, as they are added to the residual stream and constitute the majority of the hidden state’s value.

Existing self-interpretation techniques, such as Patchscopes, often yield good results in interpreting hidden states. However, further research reveals that they sometimes fail. For instance, applying these methods to the \emph{MLP and Attention outputs} almost never produces valid explanations, and in some cases, they also fail to interpret normal hidden states.

In this work, we introduce \methodFamilyName{}, a method that extends existing self-interpretation techniques, and enables the extraction of meaning from various internal representations, such as MLP outputs and hidden states, where previous methods have failed.

Our proposed method treats internal representations as instances of \emph{features as directions} (\citet{elhage2022superposition}). We suggest that in many cases where existing self-interpretation methods fail, the underlying reason is that the vectors we aim to interpret contain \emph{features that are too weak}, meaning the model does not consider them significant enough—resulting in incorrect self-interpretation. \methodFamilyName{} addresses this by \textbf{amplifying internal feature representations, significantly improving self-interpretation methods and enabling the model to successfully explain even seemingly weak features}. \emph{Figure \S\ref{fig:mainfig}} shows \methodFamilyName{}'s interpretation process of an MLP output. 

Moreover, we suggest that our method of amplification closely resembles \emph{Classifier-Free Guidance (CFG)} (\citet{ho2022classifierfreediffusionguidance}), a widely used technique in Diffusion Models.

We conduct a series of experiments (\S\ref{sec:experiments}) to evaluate the benefits and effectiveness of \methodFamilyName{} compared to prior methods, demonstrating its success across various prompts, layers, and amplifiers.

Leveraging the new capabilities introduced by \methodFamilyName{}, we developed a highly flexible framework (\S\ref{sec:application}) for interpreting different types of internal representations—including MLP outputs, the post-attention-pre-MLP residual stream, and hidden states. 

The \methodFamilyName{} framework introduces several key features that enhance interpretability by providing greater flexibility and control over internal representations inspection in language models.

One of the core capabilities of the framework is the \textbf{automatic identification of the ideal amplifier}. This feature scans multiple amplification configurations and selects the most effective one for revealing meaningful features within the model’s internal activations. By automating this process, researchers can focus on interpretation without the need for manual tuning, ensuring optimal results with minimal effort.

In addition to automatic selection, the framework allows for \textbf{manual inspection of interpretations with different amplifiers}. Researchers can explore how different amplification strengths and configurations affect interpretation results, providing deeper insights into the influence of specific activations. This feature is particularly useful for understanding how sensitive a model is to various internal signals and for validating findings across different amplifier settings.

Another powerful aspect of the framework is \textbf{the ability to patch activations into different layers of the target prompt}. Similarly to \emph{Patchscopes}, our layer-patching mechanism enables researchers to test how amplifying specific representations affects downstream computations. By injecting amplified signals into different layers, researchers can investigate how information flows through the model and how different levels of amplification contribute to final predictions.

Finally, the framework includes a \textbf{flexible selection mechanism for choosing specific tokens and layers to interpret}. Researchers can dynamically configure which layer’s outputs they wish to analyze and which token representations to focus on. This fine-grained control allows for targeted investigations into specific behaviors, making it easier to study phenomena such as token interactions, attention dynamics, and the role of hidden state transformations at different processing stages.

Together, these features make \methodFamilyName{} a powerful tool for mechanistic interpretability, providing both automation and hands-on control to uncover hidden structures in large language models.

To conclude, \methodFamilyName{}'s perspective, amplification techniques, and observations, opens up opportunities for further novel inspection techniques.

\section{Related Work}
\label{sec:related-work}

\subsection{Transformer Interpretability}
Transformer interpretability has become an increasingly prominent research focus as large language models continue to grow in complexity. Early efforts centered on analyzing attention patterns, but recent methods study deeper mechanisms within hidden states to illuminate how transformers process and store information (\citet{elhage2021mathematical}). Understanding these internal representations is crucial for explaining model predictions and mitigating undesirable behaviors. Additional techniques included training linear classifiers, called probes, on top of hidden representations (\citet{alain2017understanding, belinkov2019analysis, belinkov2022probing}), other approaches included intervening mid-computation in order to identify whether a representation is critical for certain predictions (\citet{meng2022locating,singh-etal-2020-bertnesia,wang2022interpretability, conmy2023automated,geva-etal-2023-dissecting}). 

\subsection{Vocabulary Space Projection}
A notable direction in interpretability is \emph{vocabulary space projection}, where hidden layers are mapped back to the token distribution. \emph{Logit Lens} (\citet{logitlens}) popularized this idea by inspecting intermediate logits, while \emph{Tuned Lens} (\citet{belrose2023eliciting}) refined it with learned transformations to make these projections more accurate. Variants of such approaches have shown how tracking evolving token-level distributions across layers provides insights into the gradual construction of meaning.

\subsection{Activation Patching and Self-Interpretation Methods}

Another line of work explores \emph{activation patching}, where researchers modify or swap hidden states to examine the causal role of specific activations. \emph{Patchscopes} (\citet{ghandeharioun2024patchscope}) introduced this by allowing the model to express patched-in hidden representations in human-like natural language, rather than restricting them to probability distributions, proving to be an effective method for interpretation. \emph{SeLFIE} (\citet{chen2023selfie}) further develops this approach, demonstrating its applicability across various tasks. \emph{Speaking Probes} (\citet{dar2023speakingprobes}) takes a related approach by patching model parameters—specifically, the MLP keys (sub-components of the first matrix in the MLP component)—and interpreting them in natural language.

\emph{Patchscopes} introduced the mapping function $\transformation$, which is applied to hidden states before patching them into the target prompt. As mentioned earlier, \methodFamilyName{} can be used to interpret hidden states, serving as Patchscope’s mapping function in this context. Unlike existing mapping functions, \methodFamilyName{} follows the ``features as directions'' approach, amplifying features rather than merely projecting them.

Beyond hidden states, \methodFamilyName{} also effectively interprets MLP outputs—a use case that Patchscopes and other self-interpreting methods were not originally designed for and rarely succeed in, despite its resemblance to their intended functionality.

\subsection{Features as Directions and The Superposition Hypothesis}
The \emph{features-as-directions} perspective has gained increasing attention in the field of Mechanistic Interpretability, proposing that concepts within models correspond to directions in high-dimensional embeddings (\citet{elhage2022superposition}). When multiple conceptual directions combine, they form "superposed" representations, complicating simple linear interpretations. The Superposition Hypothesis suggests that neural networks \emph{``want to represent more features than they have neurons''}, so they exploit a property of high-dimensional spaces to simulate a model with many more neurons (\emph{"Superposition"}; \citet{arora2018linearalgebraicstructureword, olah2020zoom}) - a behavior that differentiates it from classic PCA (Principal Component Analysis) approaches that rely on orthogonality.
Understanding how these directional features combine and interfere is essential for disentangling how models encode and blend various attributes in a single activation vector.
Additional significant research from \emph{Anthropic} has explored using Sparse Autoencoders to interpret models, particularly targeting MLP outputs, also viewing the vectors as composition of features (\citet{bricken2023monosemanticity} and 
\emph{features as decomposition}).
While Sparse Autoencoders yield strong results, they require extensive additional training—something \methodFamilyName{} does not.

\subsection{Classifier-Free Guidance}
In diffusion models, Classifier-Free Guidance (CFG) (\citet{ho2022classifierfreediffusionguidance}) enhances the alignment of generated outputs with a given condition (such as a text prompt). It is widely used in diffusion-based models (\citet{chen2024overviewdiffusionmodelsapplications}).

CFG operates by \emph{amplifying the direction that represents the condition}, where the difference between conditional and unconditional predictions defines this direction in the model's latent space. By amplifying this directional shift, the model is effectively \emph{steered} toward outputs that better align with the given condition. Reapplying this shift further reinforces alignment, emphasizing the semantic meaning embedded in the conditional guidance.

This technique is particularly useful in \emph{text-to-image generation} models like \emph{Stable Diffusion} and \emph{GLIDE}, as it improves adherence to prompts without requiring an external classifier.

\subsection{Diffusion Models Interpretability}
Diffusion models have also attracted significant interest from interpretability researchers. In recent years, various approaches have been proposed to analyze the latent space of diffusion models (\citet{haas2024discoveringinterpretabledirectionssemantic}).

Further work has focused on interpreting latent directions within diffusion models and leveraging these interpretations for diverse applications, such as alignment and safety (\citet{li2024selfdiscoveringinterpretablediffusionlatent}), and utilizing diffusion models in order to decode textual concepts to their main components (\citet{chefer2023hidden}).

\section{\methodFamilyName{}}
\label{sec:superscopes}
In this section, we introduce \methodFamilyName{} and demonstrate how it amplifies feature meanings in internal representations, enhancing interpretability. Additionally, we show how \methodFamilyName{} parallels techniques from diffusion models and can be viewed as their counterpart in mechanistic interpretability.

\subsection{Recap: Patchscopes}
\label{sec:subsec-patchscopes}
The core idea behind Patchscopes is to leverage the human-like generation capabilities of LLMs to \emph{"translate"} internal representations into human-readable text.

More specifically, Patchscopes extracts a hidden representation from one forward pass and \emph{patches} it into a different inference pass using a prompt that encourages the model to interpret the \emph{patched-in} representation. While Patchscopes supports cross-model patching, in \methodFamilyName{}, we focus on same-model patching—where the hidden state is injected into a forward pass of the same model but with a different input.

Formally, given an input sequence of $\sourceSeqLen$ tokens $\sourcePrompt = \langle s_1, ..., s_{\sourceSeqLen} \rangle$ and a model $\sourceModel$ with $\numSourceLayers$ layers, $\sourceHidden_{\sourcePosition}^{\sourceLayer}$ denotes the hidden representation obtained at layer $\sourceLayer \in [1, \ldots, \numSourceLayers]$ and position $\sourcePosition \in [1, \ldots, \sourceSeqLen]$, when running $\sourceModel$ on $\sourcePrompt$.

In order to extract the meaning of $\sourceHidden_{\sourcePosition}^{\sourceLayer}$, we consider a separate inference pass of the model $\sourceModel$ on a target sequence  $\targetPrompt = \langle t_1, \ldots, t_{\targetSeqLen} \rangle$ of $\targetSeqLen$ tokens. 
 (\emph{Patchscopes allows using a different model as the target model. Recall, we focus on same-model patching}).

The target sequence is designed to encourage the model to \emph{"translate"} the token at position ${\targetPosition} \in [1, \ldots, \targetSeqLen]$. A simple example of a target prompt is: \emph{"The meaning of X is:"}, where $\targetPosition$ corresponds to the token representing \emph{"X"} in the sentence.

To perform interpretation, Patchscopes selects a hidden state $\targetHidden_{\targetPosition}^{\targetLayer}$ at layer $\targetLayer \in [1, \ldots, \numSourceLayers]$ during the execution of $\sourceModel$ on $\targetPrompt$. During inference, it dynamically replaces $\targetHidden_{\targetPosition}^{\targetLayer}$ with $\sourceHidden_{\sourcePosition}^{\sourceLayer}$. As $\sourceModel$ continues generating tokens on $\targetPrompt$, the structure of our target prompt encourages it to produce human-readable text that translates the hidden state $\sourceHidden_{\sourcePosition}^{\sourceLayer}$ into words.

In addition, Patchscopes introduces a mapping function $\transformation({\sourceHidden}; \vtheta): \mathbb{R}^{\sourceHiddenDim} \to \mathbb{R}^{\sourceHiddenDim}$ parameterized by $\vtheta$ that operates on hidden representations of $\sourceModel$, which can be applied to $\sourceHidden_{\sourcePosition}^{\sourceLayer}$ before patching it into the target prompt. Patchscopes suggests that this function can be the identity function, a linear or affine transformation learned from task-specific representation pairs, or a more complex function incorporating additional data sources.

\subsection{Challenges of Self-Interpretation Methods for MLP Outputs vs. Hidden States}

A natural approach to take, following Patchscopes and other self-interpreting techniques, could be to apply the techniques to other types of internal representations, namely the Attention and MLP outputs. In this work we focus on MLP outputs.

Formally, we denote ${\sourceMLP}_{\sourcePosition}^{\sourceLayer}$ as the MLP output before it is added to the residual stream, where $\sourceLayer \in [1, \ldots, \numSourceLayers]$ indicates the layer from which this output is taken, and $\sourcePosition \in [1, \ldots, \sourceSeqLen]$ represents the position of the token for which this MLP output is generated in the source prompt.

Furthermore, we denote ${\sourcePreMLP}_{\sourcePosition}^{\sourceLayer} = {\sourceHidden}_{\sourcePosition}^{\sourceLayer} - {\sourceMLP}_{\sourcePosition}^{\sourceLayer}$ as the value of the residual stream before the addition of the MLP output.

The naive approach of applying Patchscopes to MLP outputs turns out to be ineffective, as the interpretation \emph{almost} never appears to be meaningfully related.

Selecting the appropriate token as the target for MLP-output interpretation follows existing approaches (\citet{meng2022locating, ghandeharioun2024patchscope}) for analyzing entity resolution. These studies suggest that the model constructs a subject representation at the final token of the entity name.

Examining the residual stream before the addition of MLP outputs (${\sourcePreMLP}_{\sourcePosition}^{\sourceLayer}$) at the final token of the entity name reveals an interesting phenomenon: applying Patchscopes to ${\sourcePreMLP}_{\sourcePosition}^{\sourceLayer}$ yields a meaningful explanation. Similarly, applying Patchscopes to the hidden state of the same layer, ${\sourceHidden}_{\sourcePosition}^{\sourceLayer}$, also produces a meaningful explanation—\textbf{one that is much more contextualized than the Pre-MLP residual stream}.

However, applying Patchscopes to the MLP output, ${\sourceMLP}_{\sourcePosition}^{\sourceLayer}$, results in a \textbf{meaningless interpretation}—a \emph{rather odd outcome}, given that we know this MLP output has altered ${\sourcePreMLP}_{\sourcePosition}^{\sourceLayer}$ in a way that \textbf{changes its meaning}. See (\S\ref{sec:experiments}) for further analysis.

This phenomenon, where the MLP output clearly influences the hidden state’s meaning yet lacks an interpretable meaning on its own, is noteworthy. In this work, we analyze this behavior in depth, explore its implications, and demonstrate how to interpret the MLP output itself.

\subsection{Features as Directions and Superpositions}
To examine the hypothesis underlying MLP output interpretation through self-interpretation methods, we first recap the concept of \emph{``features as directions''} and the idea behind \emph{``The Superposition Hypothesis''} (\citet{elhage2022superposition, arora2018linearalgebraicstructureword, olah2020zoom}). The core concept behind \emph{``features as directions''} is that features of hidden states are represented as \textbf{directions in activation space}. Although this is not a trivial claim, research has made significant progress in supporting it. 

Several major results and approaches reinforce this idea. A particularly notable one is vector arithmetic on word embeddings. Many are already familiar with the famous work of \citet{mikolov2013linguistic}, which demonstrates that vector arithmetic applies to word embeddings. Specifically, the following identity holds (see also \citet{levy2014linguistic}):

$V("King")-V("Man")+V("Woman")=V("Queen")$.

Some other major works include: similar \emph{"vector arithmetic"} and interpretable direction results found in generative adversarial networks (\citet{radford2016unsupervised}); a significant body of research identifying neurons with \emph{interpretable behavior} in RNNs (\citet{karpathy2015visualizing}), CNNs (\citet{zhou2015objectdetectors}), and GANs (\citet{bau2020understanding}).

Additionally, other approaches explore the concept of \emph{universality}—the idea that analogous neurons responding to the same properties can be found across different networks (\citet{schubert2021highlow})—alongside many more findings in the field.

The concept of \emph{Superposition} suggests that models encode more features than available dimensions, forcing some concepts and features to overlap or \emph{interfere} with one another.

\subsection{\methodFamilyName{}: Amplifying Weak Features}
\label{subsec:amp_weak_features}
Given that different internal representations can be viewed as instances of \textbf{features as directions} and superposition theory, our work suggests that the \emph{failed self-interpretation of MLP outputs} can be explained by the following hypothesis: MLP outputs encode changes that are exceedingly weak, making it impossible for the model to translate them on their own.

To address this, we propose amplifying the effect of these features, thereby \textbf{increasing the significance of each direction}, meaning that each feature carries more weight.

Formally, we define the amplification of the MLP outputs we aim to interpret as:

${\targetMLP}_{\sourcePosition}^{\sourceLayer} = \alpha \cdot {\sourceMLP}_{\sourcePosition}^{\sourceLayer}$

Using Patchscopes now involves replacing ${\targetHidden}_{\targetPosition}^{\targetLayer}$ with ${\targetMLP}_{\sourcePosition}^{\sourceLayer}$, the amplified MLP output. Our results demonstrate that \textbf{amplifying internal representations produces interpretable meanings}. See \S\ref{sec:experiments} for further analysis.

Similarly, applying amplification to certain hidden states that Patchscopes fails to interpret also produces interpretable meanings. Formally, this means that we replace ${\targetHidden}_{\targetPosition}^{\targetLayer}$ with $\alpha \cdot {\sourceHidden}_{\sourcePosition}^{\sourceLayer}$.

The selection of $\alpha$ parallels the choice of the guidance scale in Classifier-Free Guidance (CFG)—a larger $\alpha$ enhances feature emphasis. However, setting $\alpha$ too high may distort the vector, leading to less interpretable and distorted results. The \methodFamilyName{} framework automatically determines $\alpha$ values based on cosine similarity. A detailed analysis of our scaling approach is provided in \S\ref{sec:experiments}.

\subsection{\methodFamilyName{} as a variation of Classifier-Free Guidance (CFG)}
Classifier-Free Guidance (CFG) (\citet{ho2022classifierfreediffusionguidance}) is a widely used technique in diffusion models that improves the alignment of generated samples with a given condition, such as a text prompt in text-to-image generation. It has been broadly adopted in diffusion-based models (\citet{chen2024overviewdiffusionmodelsapplications}).

CFG operates by first computing an unconditional output, followed by a conditioned one. It then determines the difference between the conditional and unconditional outputs, defining a direction in the model's latent space. By amplifying this direction, the model is effectively \emph{steered} toward outputs that better align with the given condition. Increasing the scale factor further reinforces alignment, emphasizing the semantic meaning embedded in the conditional guidance.

The CFG amplification is expressed as:
\[
\epsilon = \epsilon_{\text{uncond}} + w \cdot (\epsilon_{\text{cond}} - \epsilon_{\text{uncond}})
\]

where:
\begin{itemize}
    \item \( \epsilon_{\text{uncond}} \) represents the model's noise prediction when no conditioning information is provided.
    \item \( \epsilon_{\text{cond}} \) represents the noise prediction when conditioning information (e.g., a text prompt) is included.
    \item \( w \) is the \emph{guidance scale}, which controls the strength of conditioning in the generated output.
\end{itemize}

The equation essentially performs an \emph{amplified directional shift} in prediction space, pushing the generated output closer to the conditioned estimate. By adjusting \( w \):
\begin{itemize}
    \item If \( w = 0 \), the model generates samples \emph{unconditionally}, without any reliance on the condition. \
    \item If \( w = 1 \), the model follows the \emph{standard conditional} generation process.
    \item If \( w > 1 \), the model \emph{exaggerates} the influence of the condition, leading to more precise alignment with the input prompt.
\end{itemize}

The core idea is that when given a condition (e.g., a text prompt like \emph{"blue fluffy dog"}) and comparing the output to an unconditioned one, subtracting the two produces a direction that represents \emph{"blue fluffy dog"}.

By repeatedly adding this direction (controlled by the guidance scale factor), the model emphasizes the condition more strongly, resulting in an output that better aligns with the given prompt.

\begin{figure} [h]
    \centering
    \includegraphics[width=0.5\textwidth]{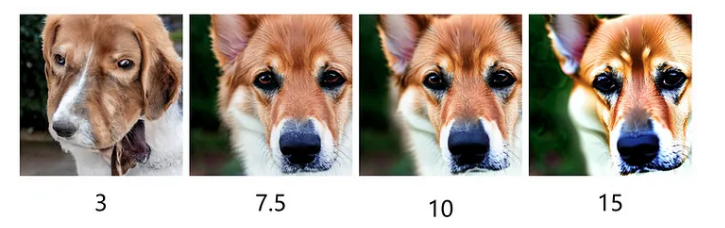} 
    \caption{\emph{An example of Stable Diffusion with the text guidance "dog" and varying CFG guidance scales.}}
    \label{fig:dogCFGexample}
\end{figure}

As shown in Figure \ref{fig:dogCFGexample}, a higher guidance scale strengthens the emphasis on the "dog." However, an excessively high guidance scale is also suboptimal, as demonstrated in the figure.

A similar example from \methodFamilyName{} is presented below, where we examine the amplification of the MLP output for the token \emph{"ppers"} in the prompt \emph{"Red Hot Chili Peppers"}:

Interpreting it with a low amplification factor yields a nonsensical interpretation:\newline
$Superscopes(\sourceMLP, \alpha=3, l=5) = $ "Chemical Compound" 

Increasing the amplification factor allows it to capture more meaning:\newline
$Superscopes(\sourceMLP, \alpha=9, l=5) = $"Rock band"

Further increasing it encapsulates even more meaning:\newline
$Superscopes(\sourceMLP, \alpha=12, l=5) = $"Band from California"

Examining this further, \textbf{we observe a clear similarity between CFG and \methodFamilyName{}} particularly in the interpretation of MLP outputs. MLP outputs play a significant role in shaping meaning, analogous to the directional shift induced by guidance in diffusion models. This behavior is notably similar, as increasing the scaling factor further emphasizes the features of the amplified vector.

We also observe a formal similarity. As previously mentioned, Patchscopes commonly succeeds in interpreting ${\sourcePreMLP}_{\sourcePosition}^{\sourceLayer}$ and ${\sourceHidden}_{\sourcePosition}^{\sourceLayer}$ but fails to interpret ${\sourceMLP}_{\sourcePosition}^{\sourceLayer}$.

Thus, we define ${\sourceMLP}_{\sourcePosition}^{\sourceLayer}$ as:

${\sourceMLP}_{\sourcePosition}^{\sourceLayer} = {\sourceHidden}_{\sourcePosition}^{\sourceLayer} - {\sourcePreMLP}_{\sourcePosition}^{\sourceLayer}$

This formulation essentially views ${\sourceMLP}_{\sourcePosition}^{\sourceLayer}$ as the \emph{"directional shift"} needed to move from ${\sourcePreMLP}_{\sourcePosition}^{\sourceLayer}$ to get to ${\sourceHidden}_{\sourcePosition}^{\sourceLayer}$.

Conceptually, this closely resembles the directional shift used in CFG:

$\epsilon_{\text{cond}} - \epsilon_{\text{uncond}}$

This resemblance highlights both the logical and formal similarity between \methodFamilyName{} and Classifier-Free Guidance, as both amplify meaning by scaling a directional shift in a similar manner.

\section{Experiments}
\label{sec:experiments}
\begin{figure}[h]
    \centering
    \includegraphics[width=0.49\textwidth]{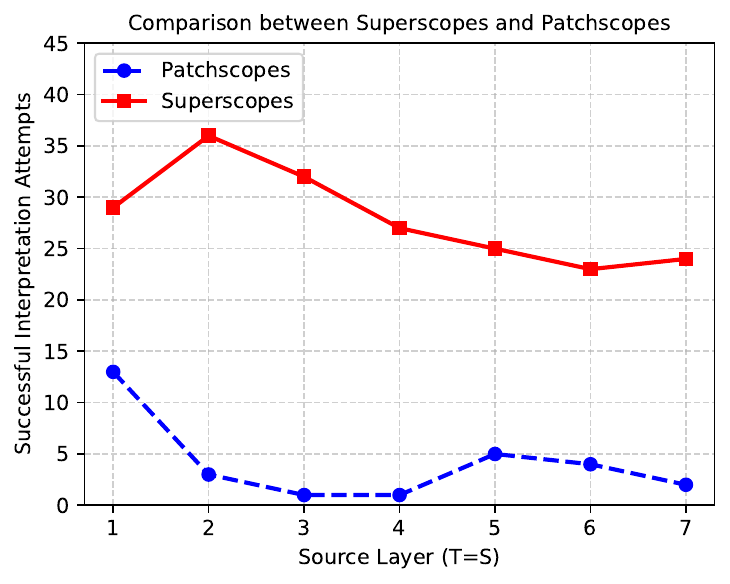}
     \captionof{figure}{A graph measuring the amount of successful interpretations of MLP outputs, by \methodFamilyName{} and \emph{Patchscopes}, over different layers. See \S\ref{subsec:amp_mlp_outputs} for further analysis.}
    \label{fig:barfig}
\end{figure}
\begin{figure*} [t]
\centering
\footnotesize 
\begin{tabular}{%
  >{\centering\arraybackslash}p{0.8cm}  
  >{\arraybackslash}p{3.0cm}            
  >{\arraybackslash}p{3.0cm}  
  >{\centering\arraybackslash}p{0.8cm}            
  >{\arraybackslash}p{3.0cm}  
  >{\centering\arraybackslash}p{0.8cm}
}
\toprule
\textbf{Layer} &
\textbf{Original Hidden State Interpretation} &
\textbf{Original MLP Output Interpretation} &
\textbf{\qmark/\xmark} &
\textbf{Amplified MLP Output Interpretation} &
\textbf{\qmark/\xmark}\\
\midrule
2 &
\textit{(Live) Aid: A charity single recorded by..} &
\textit{shortened form of the word "privile...} (incorrect interpretation) &
\xmark &
\textit{\textbf{A live television variety show}} ($\alpha$ amplifier = 15) &
\qmark \\
3 &
\textbf{American sketch comedy and variety show} &
 \centering{--}&
 --&
 \centering{--}&
 --
 \\
\bottomrule
\end{tabular}
\captionof{table}{An example of applying \emph{Patchscopes} to the token \textbf{“Live”} from the prompt \textbf{“Saturday Night Live”} at layer 2, before hidden state \emph{contextualization} occurs (first observed at layer 3). The raw MLP output appears nonsensical; however, once amplified, it correctly reflects the contextual meaning of the sentence.}
\label{tab:mlp_interpretability_example}
\end{figure*}

\begin{figure*} [t]
\centering
\footnotesize 
\begin{tabular}{%
  >{\centering\arraybackslash}p{0.8cm}  
  >{\arraybackslash}p{3.0cm}            
  >{\arraybackslash}p{3.0cm}  
  >{\centering\arraybackslash}p{0.8cm}            
  >{\arraybackslash}p{3.0cm}  
  >{\centering\arraybackslash}p{0.8cm}
}

\midrule
1 &
\textit{Ahead of one's time} &
\textit{Time period after the present} (incorrect interpretation) &
\xmark &
\textit{\textbf{1985 film, Mart...}} ($\alpha$ amplifier = 9) &
\qmark \\
2 &
\textbf{Science fiction film trilogy, and the ...} &
 \centering{--}&
 --&
 \centering{--}&
 --
 \\
\bottomrule
\end{tabular}
\captionof{table}{Another example of MLP output interpretation. This example applies \emph{Patchscopes} to the token \textbf{“Future”} from the prompt \textbf{“Back to the Future”} at layer 1, before hidden state \emph{contextualization} occurs (first observed at layer 2). The raw MLP output relates to "Future" but lacks contextualized meaning. However, once amplified, it correctly reflects the contextual meaning of the sentence—referring to the 1985 science fiction movie Back to the Future, starring \emph{Marty McFly}.}
\label{tab:mlp_interpretability_example2}
\end{figure*}
\begin{figure*} [h!]
\centering
\footnotesize 
\begin{tabular}{%
  >{\centering\arraybackslash}p{0.8cm}  
  >{\arraybackslash}p{3.0cm}            
  >{\arraybackslash}p{0.8cm}  
  >{\arraybackslash}p{3.0cm}  
  >{\centering\arraybackslash}p{0.8cm}            
}
\toprule
\textbf{Layer} &
\textbf{Original Hidden State Interpretation} &
\textbf{\qmark/\xmark} &
\textbf{Amplified Hidden State Interpretation} &
\textbf{\qmark/\xmark} \\
\midrule
4 &
\textit{Barack Obama: 44th President} (incorrect interpretation) &
\xmark &
\textbf{Ancient Greek king} ($\alpha$ amplifier = 3) &
\qmark \\
5 &
\textit{Barack Obama: American politician (incorrect interpretation)} &
\xmark &
\textbf{Ancient Greek king of Macedon} ($\alpha$ amplifier = 3) &
\qmark \\
6 &
\textbf{Ancient Greek king of Macedon} & 
\qmark &
\centering{--}&
--   \\
\bottomrule
\end{tabular}
\captionof{table}{An example of applying \emph{Patchscopes} to the token \textbf{“Great”} from the prompt \textbf{“Alexander the Great”} at layers 4, yields "Barack Obama", while the \emph{Superscoped} interpretation yields a correct interpretation (\emph{"Ancient Greek King"}). Similarly, at layer 5, \emph{Patchscopes} yields a "Barack Obama" while \emph{Superscopes} yields an even more precise interpretation(\emph{"of Macedon"}).}
\label{tab:hs_interpretability_example}
\end{figure*}

This section presents an evaluation of the \methodFamilyName{} technique by examining the degree of contextualization in MLP outputs (\S\ref{subsec:amp_mlp_outputs}) and hidden state representations (\S\ref{subsec:amp_hs_reps}) when these components are amplified, as compared to the original internal representation.

To further substantiate the versatility of the \methodFamilyName{} technique, we patched the amplified internal representations into various target layers. Specifically, patches were applied to both the initial layer ($\targetLayer = 0$) and the originating layer of the internal representation we wish to interpret ($\sourceLayer = \targetLayer$).

Note that we use \emph{Patchscopes} as our self-interpretation method of choice, but similar approaches, such as \emph{SelfIE}, can also be used.

\subsection{Amplifying MLP Outputs}
\label{subsec:amp_mlp_outputs}
In this experiment, we demonstrate that amplified MLP outputs, as interpreted through \emph{Patchscopes}, exhibit meaningful contextualization even when the original MLP output yields nonsensical results. Notably, we show that in some cases, these amplified outputs carry contextualized meaning \textbf{before the first layer} where the hidden state \textbf{encodes the resolved entity}.

\indent\textbf{Methods}\hspace{0.5em} We evaluated various amplification levels of MLP outputs, including a non-amplified baseline, to identify the most interpretable vectors using \emph{Patchscopes}.\newline As detailed in \S\ref{subsec:amp_weak_features}, let $\sourceLayer_c$ denote the first layer where the hidden state encodes the resolved entity (\emph{as identified by Patchscopes}); We interpret ${\targetMLP}^{\sourceLayer}_{\sourcePosition}$ starting from ${\sourceLayer=\sourceLayer_c-1}$. To determine an optimal amplification value, we conducted our tests using $\alpha=\{1,3,6,9,12,15\}$, in addition to testing intermediate values between them.\newline
To derive $\sourceLayer_c$, we followed the methodology from \textit{Patchscopes}, analyzing how large language models contextualize entity names. This involved crafting a target prompt to generate a subject description and applying it to the hidden representation at \emph{the last subject position} in the source prompt—where the model forms the subject representation (\citet{geva-etal-2023-dissecting, hernandez2023measuring}). This approach allows us to observe how the model describes the subject at each layer and identify the first layer where contextualization occurs.

\indent\textbf{Measurement Metrics}\hspace{0.5em} To measure the performance of \methodFamilyName{} on MLP outputs, we designed two distinct experiments. In the first experiment, we sought manual results that emphasize the interpretability of amplified MLP outputs, specifically examining how, in certain prompts, nearly the entire context of a sentence can be inferred from an amplified MLP output prior to hidden state contextualization.
In the second experiment, we aimed to demonstrate the superiority of Superscopes over Patchscopes using a larger set of examples. We applied Superscopes to each prompt for $l=\{1 \dots 7\}$ with the previously defined $\alpha$ values, where $\alpha=1$ corresponds to Patchscopes. Performance was evaluated using \emph{all-MiniLM-L6-v2}, a model trained for semantic similarity and sentence representations. When computing the cosine similarity score, we set a threshold of 0.3 as a reasonable indicator of success, based on empirical observations. While this value works well in our experiments, other thresholds can be chosen to achieve similar results.

\indent\textbf{Results}\hspace{0.5em} Using a set of short prompts (\emph{"Diana, Princess of Wales", "Back to the Future", "Saturday Night Live"...}), we demonstrate that \methodFamilyName{} enables the interpretation of MLP outputs that initially appear nonsensical before amplification. Table \S\ref{tab:mlp_interpretability_example} and table \S \ref{tab:mlp_interpretability_example2} illustrate two examples of a hidden state in a layer before contextualization, while also showing how the fully contextualized meaning is successfully extracted from the \emph{amplified MLP output at the same layer}. Additionally, the illustration on the front page (Figure \S\ref{fig:mainfig}) and the Superscopes application (\S\ref{sec:application}) serve as further examples of successful \methodFamilyName{} interpretations.
Through the second experiment, we demonstrated significantly superior results (see \S\ref{fig:barfig}), thereby validating the effectiveness of the \methodFamilyName{} method.

\subsection{Amplifying Hidden States Representations}
\label{subsec:amp_hs_reps}
Using the \methodFamilyName{} on hidden states representations also yields exciting results. 

Patchscopes (\S \ref{sec:subsec-patchscopes}) also introduced a mapping function, $f$, and suggests that this function can be the identity function, a linear or affine transformation learned from task-specific representation pairs, or a more complex function incorporating additional data sources. We suggest that this function can also simply be the amplification of hidden states, as introduced by \methodFamilyName{}.

In addition to successfully interpreting MLP outputs, \methodFamilyName{} also improves the ability to interpret hidden representations, which we examine in this section.

\indent\textbf{Methods}\hspace{0.5em} Similarly to the MLP output evaluation, we evaluated various amplification levels of hidden states, including a non-amplified baseline, to identify the most interpretable vectors using \emph{Patchscopes}.\newline As detailed in \S\ref{subsec:amp_weak_features}, let $\sourceLayer_c$ denote the first layer where the hidden state encodes the resolved entity (\emph{as identified by Patchscopes}); We interpret $\sourceHidden_\sourcePosition^\sourceLayer$, starting from ${\sourceLayer=\sourceLayer_c-1}$ and going backwards in the layers. To identify the optimal amplification value, we conducted our tests across a wider range of $\alpha$ values, highlighting the precision necessary for effectiveness in hidden state 
\methodFamilyName{} interpretation.\newline
To derive $\sourceLayer_c$, we followed the same methodology from the MLP output evaluation (\S \ref{subsec:amp_mlp_outputs}), by crafting a prompt that encourages the model to "translate" a hidden representation (see \S \ref{sec:subsec-patchscopes} further details).

\indent\textbf{Measurement Metrics}\hspace{0.5em} To evaluate the performance of \methodFamilyName{} on hidden state representations, we examined various short prompts and attempted to interpret hidden state representations from earlier layers than the ones interpreted with broader context through \emph{Patchscopes}. A successful attempt was defined as one in which we were able to interpret layers preceding the contextualized layer analyzed using \emph{Patchscopes}.

\indent\textbf{Results}\hspace{0.5em} Similarly to our evaluation of the MLP output amplification, using a set of short prompts (\emph{"Alexander The Great", "Red Hot Chili Peppers", "Florence and the Machine"}), we demonstrate that \methodFamilyName{} also enables the interpretation of hidden states that previously appear nonsensical before amplification. Table \S\ref{tab:hs_interpretability_example} presents an example of hidden states representations successfully interpreted using \methodFamilyName{} two layers prior to contextualization.

\section{Applications}
\label{sec:application}
\subsection{The \methodFamilyName{} Framework}
\label{sec:subsec-app-framework}
Building on our findings, we recognize the potential of using amplification for early exits and improving the understanding of Language Models' thought process and contextualization. To support this, we introduce \methodFamilyName{}: an all-in-one, fully \textbf{open-source} utility designed to evaluate outcomes across various amplification levels, target layers, and prompts

\indent\textbf{Functionality and Features}\hspace{0.5em} The application specifically focuses on the pre-MLP residual stream (also referred to as the \emph{"Post-Attention-Pre-MLP residual stream"}), MLP outputs, and final hidden-state values. Earlier, we demonstrated (\S\ref{subsec:amp_mlp_outputs} and \S\ref{subsec:amp_hs_reps}) that amplified MLP outputs and hidden representations can be interpreted more effectively when \emph{Superscoped}. To facilitate this, we developed the \methodFamilyName{} framework to identify and "fine-tune" such cases.

\methodFamilyName{} allows adjustment of the following parameters:
\begin{enumerate}
\item Inspected prompt – Fully adjustable and configurable.
\item Inspected token – Any input token can be examined using \methodFamilyName{}.
\item Source layer range – Fully adjustable and configurable.
\item Target layer – Can be configured between same-layer patching and patch-to-layer-zero configurations.
\item Automatic amplifier detection – Using the methods described in \S\ref{subsec:amp_mlp_outputs} (based on cosine similarity scores), \methodFamilyName{} automatically identifies the best amplifier for each scenario while also displaying all tested amplifiers.
\end{enumerate}

\begin{figure}[h]
\centering
\includegraphics[width=0.5\textwidth]{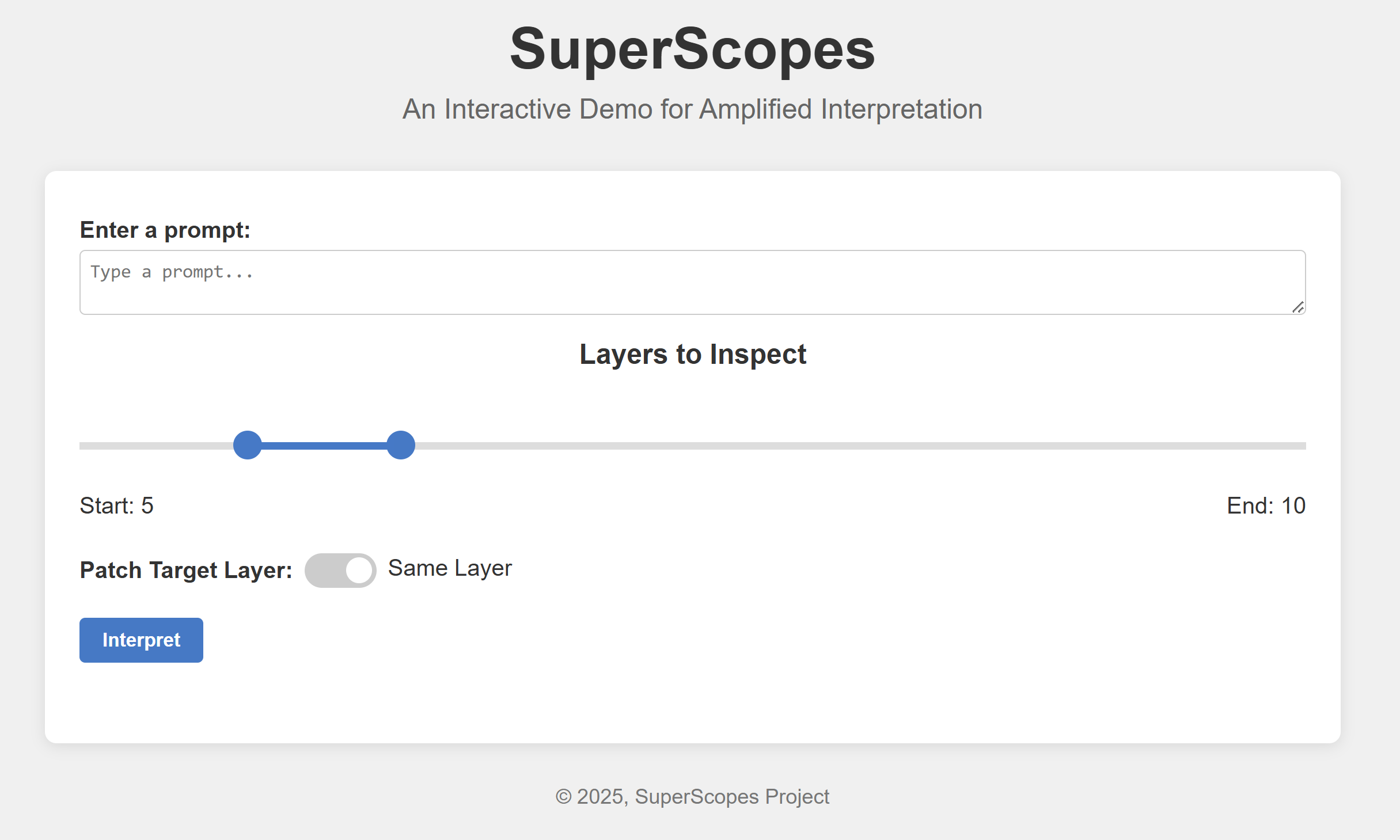}
\caption{The \methodFamilyName{} Framework}
\label{fig:superscopes_framework}
\end{figure}

\indent\textbf{Key Examples}\hspace{0.5em}
Leveraging the flexibility of our framework, we uncovered several key observations. One notable finding emerged for the source prompt \emph{"Diana, Princess of Wales"}.

\begin{figure}[h]
\centering
\includegraphics[width=0.5\textwidth]{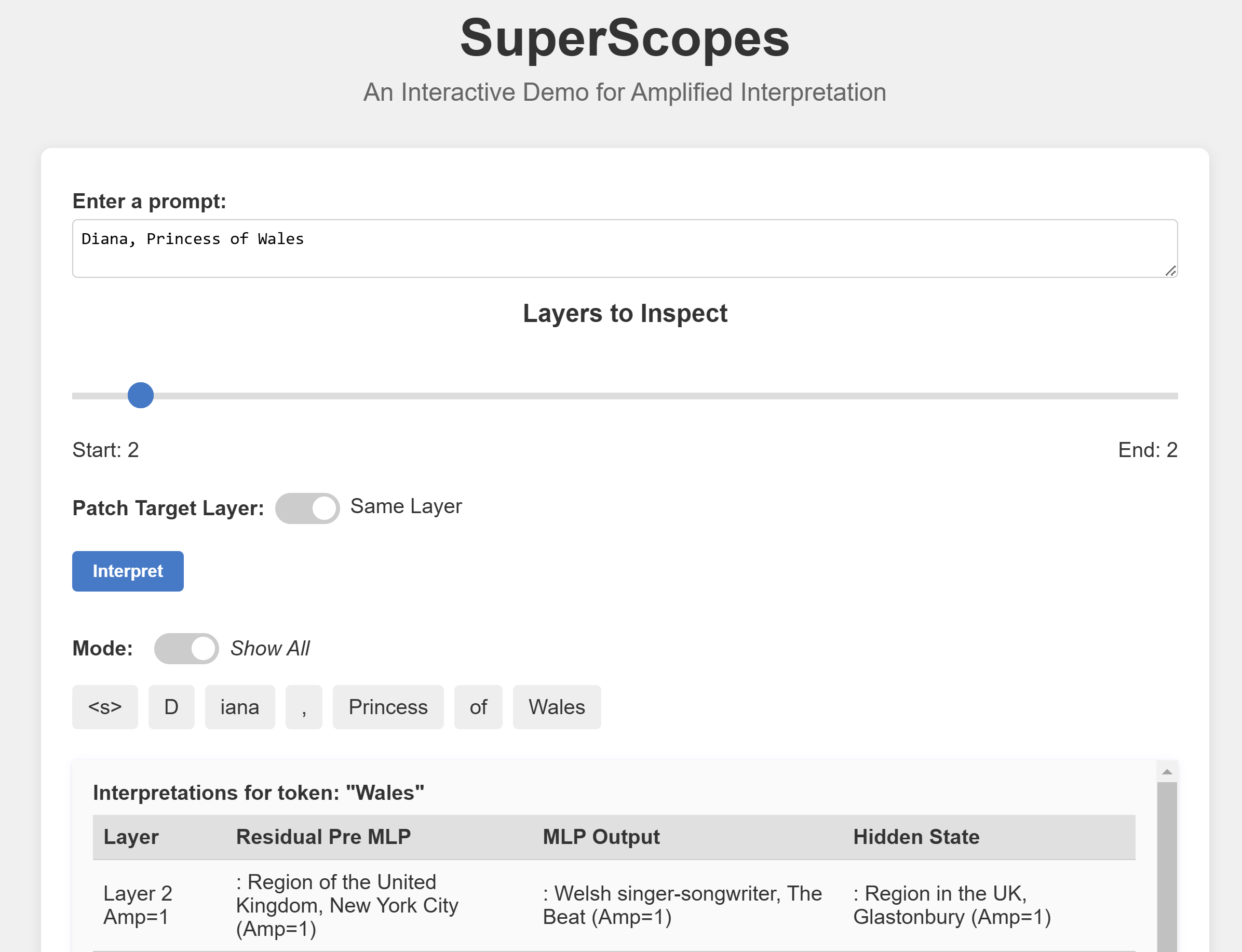}
\caption{Using an amplifier of $\alpha = 1$ (Patchscopes), we obtain partial \& incorrect interpretations for both the Pre-MLP residual and the MLP output.}
\label{fig:patchscopes_sample}
\end{figure}

\begin{figure}[h]
\centering
\includegraphics[width=0.5\textwidth]{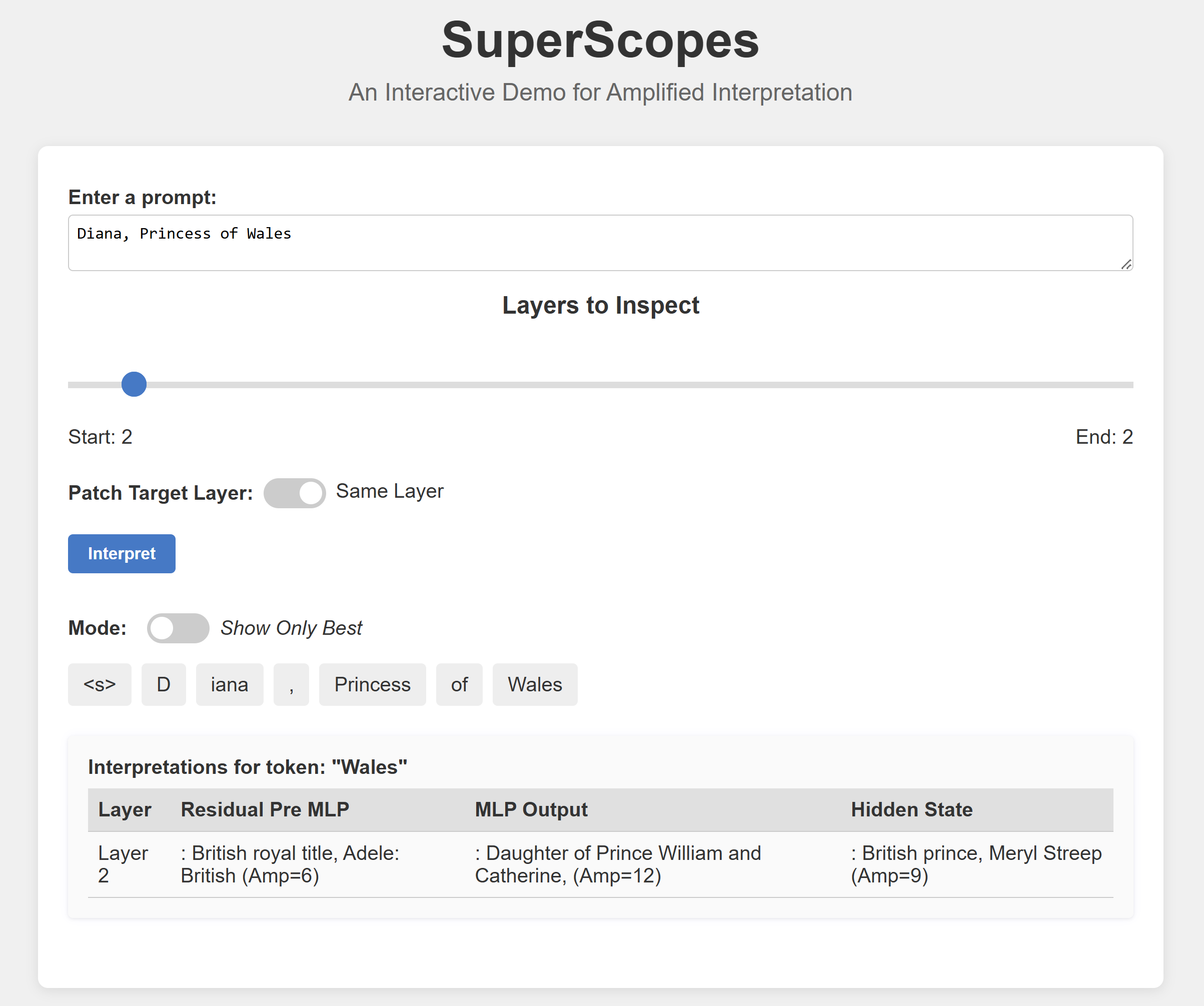}
\caption{Using \methodFamilyName{}'s automatic amplifier detection, we obtain correct interpretations for both the Pre-MLP residual and the MLP output.}
\label{fig:superscopes_sample}
\end{figure}

By using the "Show Best Results" option compared to standard \emph{Patchscopes}, we found that \methodFamilyName{} successfully revealed (\S\ref{fig:superscopes_sample}) meaning for both the hidden state and the MLP output—interpretations that were otherwise inaccessible (\S\ref{fig:patchscopes_sample}).

The \methodFamilyName{} framework provides an accessible platform for testing various amplification levels, target layers, and prompts, enabling a deeper exploration of their impact on different interpretations.

\section{Conclusion and Future Work}
In this paper, we introduced \methodFamilyName{}, a method that extends Patchscopes by amplifying hidden states and MLP outputs before patching. We demonstrated that even hidden states previously considered uninterpretable (e.g., MLP residual outputs in large Transformer layers) can reveal coherent semantic content when appropriately scaled.

Our results reinforce the concept of \emph{features as directions} in large language models, highlighting that interpretability can be improved by controlling the magnitude of these directions. Furthermore, we show that \methodFamilyName{} can also be understood as a variant of Classifier-Free Guidance (CFG), exhibiting similar behaviors in internal representations.

We also introduce the \methodFamilyName{} Framework to facilitate further research and exploration of the effects of amplification on different components of a model. 

Further research directions include extracting meaning from attention outputs, which initially appear to exhibit a "denser form" of superposition (partially due to attention heads). Another avenue for future work is refining the \methodFamilyName{} methodology—one possibility is developing a more effective approach for selecting the optimal amplifier, while others involve exploring alternative ways to amplify meaning within \methodFamilyName{}.

Additionally, similar to Tuned Lens (\citet{belrose2023eliciting}), there may be a more suitable "base" vector than directly patching the MLP output. Other directions for future work include searching for better ways to patch different internal representations (e.g., MLP outputs). Our work also explored adding MLP outputs across multiple layers in a single inference pass; however, future research may uncover new methods to make this approach more effective.

\section{Acknowledgments}
We would like to thank Guy Dar, Shahar Satamkar, Hila Chefer, Yossi Gandelsman and Dr. Eli David for their valuable feedback and support throughout our work.

\vspace{1cm} The code for this project is available on GitHub at \href{https://github.com/GalNivs/Superscopes}{Superscopes}.

\end{document}